\documentclass[10pt,twocolumn,letterpaper]{article}

\usepackage{cvpr}              

\usepackage{graphicx}
\usepackage{amsmath}
\usepackage{amssymb}
\usepackage{booktabs}
\graphicspath{{images/}}

\usepackage[pagebackref,breaklinks,colorlinks]{hyperref}

\usepackage[capitalize]{cleveref}
\crefname{section}{Sec.}{Secs.}
\Crefname{section}{Section}{Sections}
\Crefname{table}{Table}{Tables}
\crefname{table}{Tab.}{Tabs.}

\def\cvprPaperID{} 
\def\confName{CVPR}
\def\confYear{2022}

\begin{document}

\title{Semantically-aware Mask CycleGAN for Translating Artistic Portraits to Photo-realistic Visualizations}

\author{YIN Zhuohao\\
Hong Kong University of Science and Technology, Hong Kong\\
{\tt\small zyinad@connect.ust.hk}
}
\maketitle

\begin{abstract}
   Image-to-image translation (I2I) is defined as a computer vision task where the aim is to transfer images in a source domain to a target domain with minimal loss or alteration of the content representations. Major progress has been made since I2I was proposed with the invention of a variety of revolutionary generative models. Among them, GAN-based models perform exceptionally well as they are mostly tailor-made for specific domains or tasks. However, few works proposed a tailor-made method for the artistic domain.  In this project, I propose the Semantic-aware Mask CycleGAN (SMCycleGAN) architecture which can translate artistic portraits to photo-realistic visualizations. This model can generate realistic human portraits by feeding the discriminators semantically masked fake samples, thus enforcing them to make discriminative decisions with partial information so that the generators can be optimized to synthesize more realistic human portraits instead of increasing the similarity of other irrelevant components, such as the background. Experiments have shown that the SMCycleGAN generate images with significantly increased realism and minimal loss of content representations.
\end{abstract}

\section{Introduction}
\label{sec:intro}

\begin{figure}[htp]
    \centering
    \includegraphics[width=0.3\textwidth]{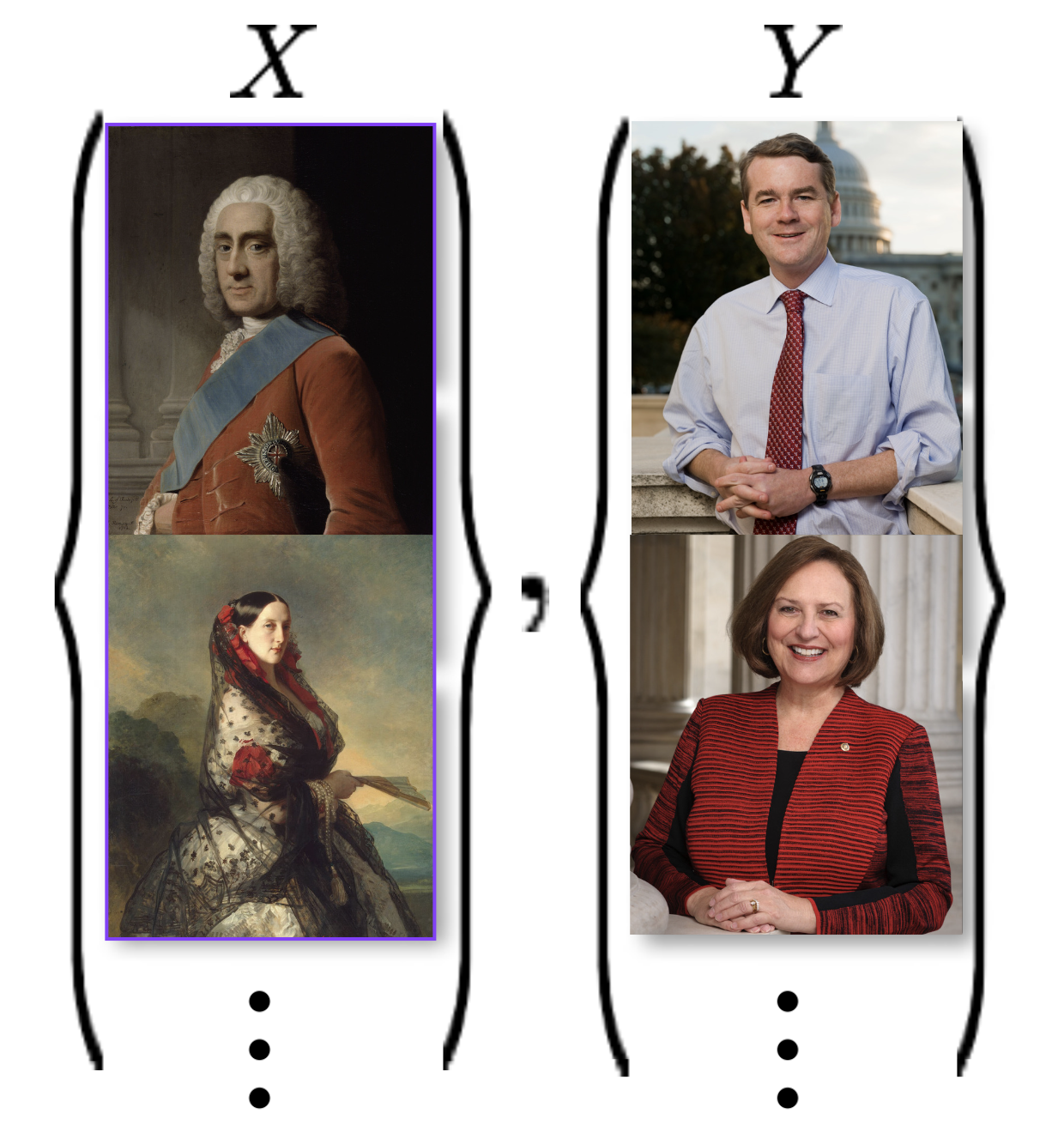}
    \caption{Unpaired data comprised of source set $X=\{ x_i\}_{i=1}^N $ and target set $Y=\{ y_i\}_{i=1}^N $.}
    \label{fig:problem statement}
\end{figure}

In 1816, the world's first camera was invented and cameras later became the dominant tool for human beings to capture moments in their lives. However, prior to this revolutionary invention, paintings served as the only feasible approach to recording visual information. Realistic as some genres of paintings may seem, such as mannerism and classicism, there still existed a gap between what was on the canvas and what was before human eyes. This project seeks to fill this gap, specifically within the domain of artistic portraits, where the aim is to restore what was presented before the painters hundreds of years ago. Given a set of art portraits $X$ and a set of realistic portrait photos $Y$ (Fig. \ref{fig:problem statement}), the model is expected to learn a mapping $G$ such that given some $x_i$,
\begin{equation}
    y_i = G(x_i) \in Y.
\end{equation}

More broadly, this is a problem of unpaired image-to-image translation \cite{zhu2017unpaired, liu2017unsupervised}, where the goal is to translate images in a source domain to a target domain while preserving the content representations. Since the proposal of CycleGAN \cite{zhu2017unpaired}, a variety of variants that were built on the cycle-consistent framework have been proposed \cite{pang2021image}, tailor-made to solve different problems. The Augmented CycleGAN (AugCGAN) extended the results of CycleGAN from one-to-one mappings to many-to-many mappings by adopting latent codes to capture missing information in the generated images and learning the mappings in the augmented space\cite{almahairi2018augmented}. Thus, the AugCGAN is capable of generating diverse images when different latent codes are sampled. The Unsupervised Image-to-Image Translation Networks (UNIT) is built on the assumption that any pair of images in two domains that are intertranslatable has a shared latent code in a shared latent space. The training process of UNIT is performed in the latent space after encoding the raw inputs.

Under the unpaired I2I setting, a major drawback of the vanilla CycleGAN \cite{zhu2017unpaired} is that the mappings between two domains are learned as a whole for each image pair. That is, each image is regarded as one distribution and the generators are optimized to minimize the difference between the distributions in the source domain and those in the target domain. However, such methods may not yield satisfactory results when the input images consist of semantically different objects. For example, art portraits are generally comprised of a human subject and a background. Naturally, distinct objects tend to belong to different distributions and it is much more appropriate to view one image as a mixture of different distributions. Based on this assumption, I propose the Semantic-aware Mask CycleGAN (SMCycleGAN), which applies semantic segmentation masks on generated samples masking out the background before they are fed to the discriminators to differentiate. This enforces the discriminators to make discriminative decisions based on partial information. Consequently, the generators can be optimized particularly to synthesize more realistic human subjects rather than attempting to match the background patterns to increase realism.

My contributions are as follows. (i) I introduce the Semantic-aware Mask CycleGAN for translating artistic portraits to photo-realistic visualizations. (ii) I show that my model can learn the mapping from the artistic domain to the realistic domain for portraits and generate portraits with high realism. (iii) I show that the SMCycleGAN achieves better qualitative and quantitative results than several baseline models.

\section{Related Work}
\label{sec:related work}
\textbf{Cycle Consistency.} One of the revolutionary concept proposed by Zhu et al. \cite{zhu2017unpaired} was the cycle-consistency loss. Suppose we want to transform images between domain $X$ and domain $Y$, and we have a translator $G:X\rightarrow Y$ that receives images from domain $X$ and output their corresponding representation in domain $Y$ and another translator $F:Y\rightarrow X$ that does the opposite. Consider the translated result $G(x)$ of an image $x\in X$ that is transferred to domain $Y$ through generator $G$. Once we obtain $G(x)$, an intuitive process is that if we send $G(x)\in Y$ through $F:Y\rightarrow X$, we should be able to recover the original image $x$. The cycle-consistency loss depicts the distance between the recovered image $F(G(x))$ and the original image $x$. Adding the cycle-consistency loss to the training process encourages the generation of images to be \textit{cycle consistent}. While CycleGAN has been shown to yield sensible results on a variety of tasks, such as interconverting zebras and horses, summer landscapes and winter landscapes, and Monet paintings and natural images, it remains a generic method for I2I problems as it learns the mappings between domains regarding each image as a single distribution instead of mixtures of distributions. Additionally, CycleGAN fails to utilize any structural or semantic information embedded in the data, but merely regulates the networks' behaviors through the delicately-defined objective functions.

\textbf{Patch-retrieving Mechanism.} The mechanism of retrieving patches of natural images has been widely adopted for image synthesis tasks \cite{tan2019text2scene, tseng2020retrievegan}. Intuitively, including patches from real images can significantly increase the realism of the synthesized images. Several works extended this technique for I2I tasks \cite{tomei2018monet, tomei2019art2real}. Tomei et al. \cite{tomei2018monet} built their model on vanilla CycleGAN \cite{zhu2017unpaired} by adding the patch-retrieval mechanism for transforming Monet paintings to photo-realistic images. An external memory bank was established, storing patches of real images. During the training process, each generated image is split into patches and for each generated patch, one real patch which is the most similar is retrieved from the memory bank and paired with the generated patch. The objective included the maximization of the similarity between each generated patch and its closest real patch in the external memory bank, thus pushing the generators to produce more realistic images. 

\textbf{Exploiting Semantic Information.}
Although CycleGAN\cite{zhu2017unpaired} is proven to be effective on generating sensible translation results with consistent content representations as the cycle-consistency loss greatly reduced the space of total possible mapping functions, the reduction is still inadequate in certain problem settings so that the training process is highly stochastic and the performance could not be guaranteed. In other words, there is much problem-specific information that are failed to be exploited.

To remedy the information waste, Tomei et al. \cite{tomei2019art2real} proposed Art2Real, which is a semantically-aware architecture for translating artworks to photo-realistic images. During the training process of Art2Real, semantic segmentation is first performed on the generated samples to obtain segmentation maps. For instance, a landscape painting will typically be segmented into different components such as sky, grass, tree, etc. After segmentation, patches are produced from the segmentation maps in the same way as in \cite{tomei2018monet}, but with predicted class labels. Patches that belong to the same class will again be paired up with a real patch retrieved from an external memory bank. Different from the only memory bank where patches of all types of instances are stored in \cite{tomei2018monet}, the revised method adopted a class of memory banks, storing patches of different semantics separately. For example, a patch sampled from a segmentation map of sky will search for a patch in the memory bank of sky patches and find the one with the highest similarity to pair up with. Once all generated patches are paired up with real ones, the model is optimized to further increase the similarity between each pair, thus achieving the goal of synthesizing realistic images of the same content representations.
\section{Methods}
\label{sec:methods}
\begin{figure}[htp]
    \centering
    \includegraphics[width=8cm]{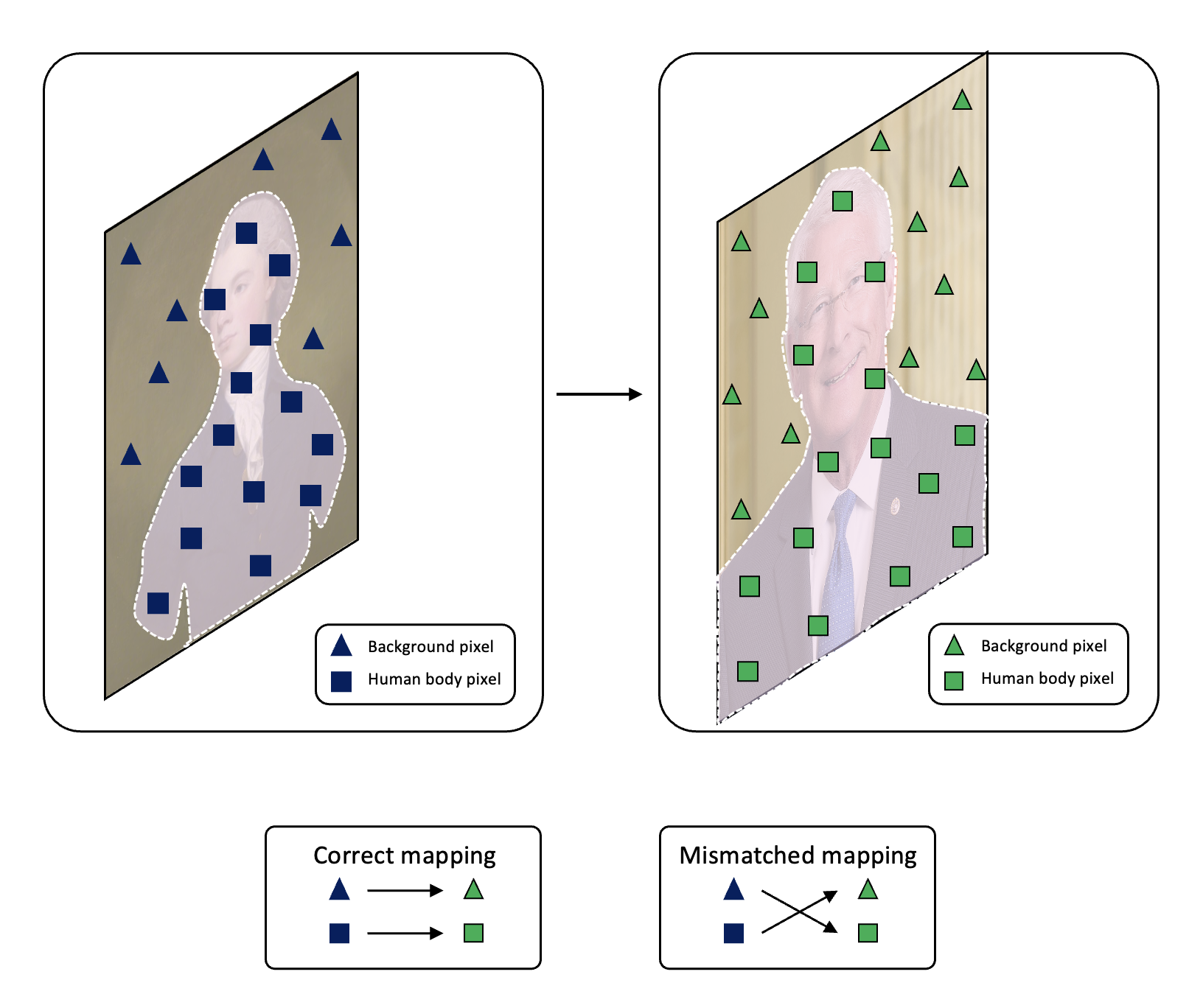}
    \caption{Different mapping patterns between two domains. The triangles represent background pixels and human subjects are represented by squares. Those in the artistic domain are colored blue and those in the realistic domain are colored green. The correct mapping pattern is shown on the left and the mismatched mapping pattern is shown on the right.}
    \label{fig:mapping}
\end{figure}

Since the goal of this project is to transfer artistic portraits to photo-realistic visualizations, my proposed approach is built upon the unpaired image-to-image translation setting and adopts the same training logic as that of the CycleGAN \cite{zhu2017unpaired}. However, the problem of this project differs from most image-to-image translation problems in that each portraits is comprised of instances that have distinct semantics, i.e. the background and the human subject. Whereas, in \cite{tomei2018monet}, they studied Monet's landscape paintings, of which the composition is much more consistent. Naturally, different semantic instances tend to have different distributions. Hence, it is only appropriate to regard each image as a mixture of distributions rather than a single distribution as a whole. Consequently, it is a must that the component distributions be learned and mapped separately. Fig. \ref{fig:mapping} illustrates the scenarios of correct mappings and mismatched mappings. In cases of mismatched mappings, even though the mixture of the two distributions are able to match up, the underlying mapping pattern is erroneous. Thus, my model introduces segmentation masks and presents the discriminators generated samples with the backgrounds ruled out, enforcing them to make discriminative decisions only judging from the human subjects. This mechanism discourages the generators from a cheating behavior that they try to increase the similarity of the backgrounds to deceive the discriminators.

\subsection{Vanilla Adversarial Loss}
\label{sec:vanilla adv}

As the core component of losses of GAN models, the adversarial loss in my model is applied to two directions of translation. Define $G:X\rightarrow Y$ to be the mapping from artistic portraits to realistic ones, and $F:Y\rightarrow X$ the opposite. Define $D_Y$ as the discriminator to spot generated realistic portraits and $D_X$ the opposite. The adversarial loss can be expressed as:
\begin{equation} 
\begin{split}
\mathcal{L}_{GAN}(G, D_Y, X, Y)=&\mathop{\mathbb{E}}_{y\in Y}[\log{D_Y(y)}]\\
        +&\mathop{\mathbb{E}}_{x\in X}[\log{(1-D_Y(G(x))}]\\
\mathcal{L}_{GAN}(F, D_X, Y, X)=&\mathop{\mathbb{E}}_{x\in X}[\log{D_X(x)}]\\
        +&\mathop{\mathbb{E}}_{y\in Y}[\log{(1-D_X(F(y))}].
\end{split}
\end{equation}

However, the vanilla adversarial loss does not tend to work well when there are semantically distinct components in the image. To this end, I introduce the adversarial loss over segmented regions in Section  \ref{sec:masked_loss}.

\subsection{Adversarial Loss over Segmentation Maps}
\label{sec:masked_loss}

\begin{figure}[htp]
    \centering
    \subfloat[]{%
        \includegraphics[width=0.23\textwidth]{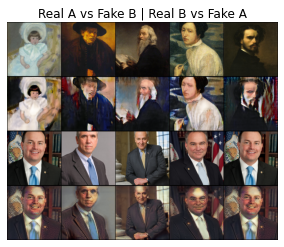}%
        \label{fig:flag1}%
        }%
    \hfill%
    \subfloat[]{%
        \includegraphics[width=0.23\textwidth]{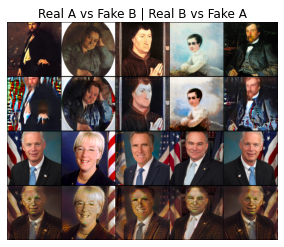}%
        \label{fig:flag2}%
        }%
    \caption{Undesirable experiment results where patterns of the flag of United States are transferred onto the generated realistic photos from the art portraits. The first row shows the original painting. The second row shows the photo-realistic visualizations of the art portraits. The third row shows the realistic portraits. The fourth row shows the artistic visualizations of the realistic portraits.}
\end{figure}

As opposed to Monet's landscape paintings studied by \cite{zhu2017unpaired} and \cite{tomei2018monet}, the artistic theme that this project aims to study is portraits, which each has a clear-cut subject. However, the vanilla adversarial loss does not distinguish between the subject and the background. Instead, it judges the authenticity of the images by a general look across the image. The drawback is that the features in the background of one domain tend to be transferred onto the subject or the background in the other domain, while neither is desired. As shown in Figure \ref{fig:flag1} and \ref{fig:flag2}, several generated images from the art portraits have blue and red stripes in the background, as well as on the subjects' clothes, which are manifest signs of the flag of the United States in the background of the realistic photos.

My proposed solution is to apply different degrees of adversarial loss on different semantically-segmented regions. Each portrait image, whether artistic or realistic, is first segmented to produce a mask that softly rules out the background by assigning low scores to background pixels and high scores to human body pixels. Denote $M_x$ to be the mask matrix for image $x$ which has the same shape as the input image. A masked image is given by:
\begin{equation}
    x_{masked} = M_x \odot x,
\end{equation}
where $\odot$ is the element-wise multiplication between M and each color channel of $x$. The adversarial loss is then applied on the masked image:
\begin{equation}
    \begin{split}
\mathcal{L}_{mGAN}(G, D_Y, X, Y)=&\mathop{\mathbb{E}}_{y\in Y}[\log{D_Y(M_y \odot y)}]\\
        +&\mathop{\mathbb{E}}_{x\in X}[\log{(1-D_Y(M_{G(x)} \odot G(x))}]\\
\mathcal{L}_{mGAN}(F, D_X, Y, X)=&\mathop{\mathbb{E}}_{x\in X}[\log{D_X(M_x \odot x)}]\\
        +&\mathop{\mathbb{E}}_{y\in Y}[\log{(1-D_X(M_{F(y)} \odot F(y))}].
\end{split}
\end{equation}

\begin{figure}[htp]
    \centering
    \includegraphics[width=0.5\textwidth]{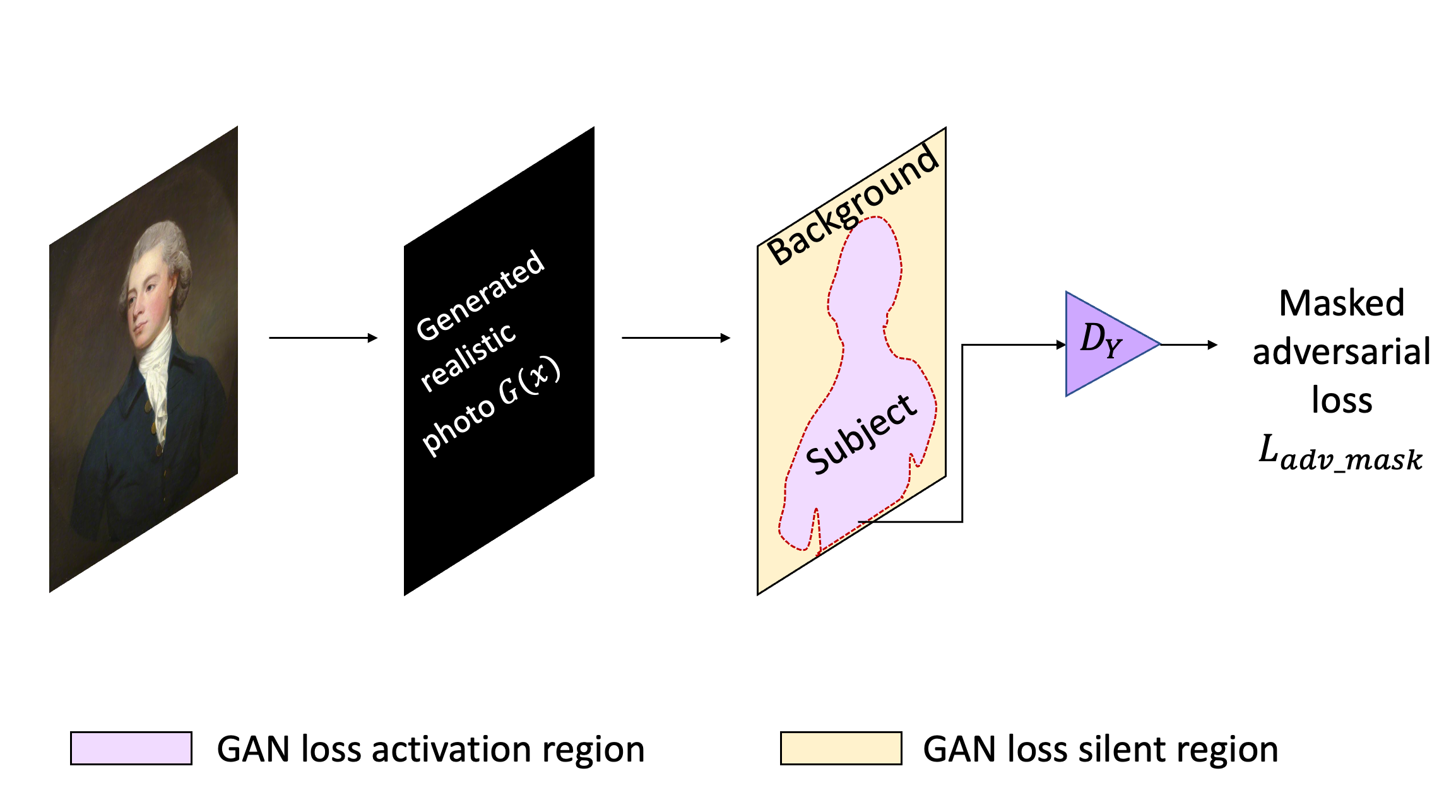}
    \caption{After translating an artistic portrait $x\in X$ to a realistic photo $G(x)\in Y$, the discriminator $D_Y$ puts its focus on the subject region of the generated image, while the background region induces a small or zero portion of the adversarial loss.}
    \label{fig:masked_loss}
\end{figure}

Fig. \ref{fig:masked_loss} demonstrates the process of computing the masked adversarial loss for one artistic portrait $x\in X$. The input first goes through the generator $G$, before the generated realistic photo is fed into a U-Net \cite{ronneberger2015u}, which segments the human body from the background and produces the aforementioned mask matrix $M_x$. $M_x$ is a matrix of the same shape as the input image $x$ and has value $1$ in the purple regions and $0$ in the yellow region in Fig. \ref{fig:masked_loss}, where the purple region is called the adversarial loss activation mask and the yellow region is called the silent region. The discriminator $D_Y$ then examines the generated realistic photo with its focus on the purple active region. Intuitively, this forces the discriminator to only look at the human body part to decide whether this sample is real or fake, which restrains the generator's behavior of learning features in the background to deceive the discriminator. 
 
\subsection{Cycle Consistency Loss}
\label{sec:cycle consistency loss}
Like many existing models tackling the unpaired image-to-image translation problem, my model also adopts the cycle consistency loss, expressed as:

\begin{equation}
\begin{split}
    \mathcal{L}_{cyc}(G, F)&=
    \mathop{\mathbb{E}}_{x\in X}[\begin{Vmatrix} F(G(x)-x) \end{Vmatrix}]\\
    &+\mathop{\mathbb{E}}_{y\in Y}[\begin{Vmatrix} G(F(y)-y) \end{Vmatrix}].
\end{split}
\end{equation}

The intuition comes from the assumption that an image after translation is expected to be restored by applying the reverse mapping to the translated image. The sequence of translations can be expressed as:
\begin{equation}
x\in X, x\rightarrow G(x) \rightarrow F(G(x)).   
\end{equation}

In theory, the reconstructed $F(G(x))$ should be identical to the original $x$. Hence, the difference between the two induces the cycle consistency loss.

\subsection{Identity Loss}
\label{sec:identity loss}

Another technique introduced by Zhu et al. \cite{zhu2017unpaired} was the identity loss. The idea is that when the input to a generator is from the target domain to begin with, the generator is encouraged to be an identity mapping that produces an identical result to the input. This regularizer has been shown particularly helpful in photo generation from paintings. Given the context of this project, it is suitable and critical to include the identity loss in the full objective.

\subsection{Full Objective}
\label{sec:full obj}
The full objective of my model is:
\begin{equation}
\begin{split}
    \mathcal{L}(G,F,D_X,D_Y)&=\mathcal{L}_{mGAN}(G,D_Y,X,Y)\\
    &+\mathcal{L}_{mGAN}(F,D_X,Y,X)\\
    &+\lambda \mathcal{L}_{cyc}(G,F)\\
    &+0.5\lambda \mathcal{L}_{identity}(G,F),
\end{split}
\end{equation}

where $\lambda$ is the importance of different components of the full objective.

\section{Experiments}
\label{sec:experiments}

\textbf{Datasets.}
Since this project involves training multiple models, multiple datasets were used to fulfill different tasks. To train the semantic segmentation model that produces segmentation masks, a subset of the \href{https://www.kaggle.com/datasets/laurentmih/aisegmentcom-matting-human-datasets}{Matting Human Dataset} on Kaggle is utilized. The dataset comes with two folders where one stores human photographs and the other stores the segmentation masks that mask out the backgrounds. All images are rescaled to $256\times 256$ pixels during both training and testing. The size of the training set is 3500. As for the artworks, they are originally retrieved from \href{https://www.wikiart.org/}{Wikiart.org} and the \href{https://www.kaggle.com/datasets/karnikakapoor/art-portraits}{Art Portraits Dataset} on Kaggle extracted the portrait category to form a new dataset. Overly abstract portraits are pruned by hand as it is unlikely to generate realistic photos from them. During the early stage of development, the \href{https://www.kaggle.com/datasets/blakefallconroy/current-us-senators?resource=download}{US Senators Dataset} on Kaggle was adopted. It is comprised of 100 portraits of the U.S. senators. Since its size is too small and the content is too domain-specific, it was deprecated and I instead used the human portrait data in the \href{https://www.kaggle.com/datasets/laurentmih/aisegmentcom-matting-human-datasets}{Matting Human Dataset} as data in the realistic domain when training the SMCycleGAN model.

\textbf{Training details.}
The segmentation model I adopted was U-Net proposed by Ronneberger et al. primarily for biomedical image segmentation \cite{ronneberger2015u}. It was trained for 15 epochs using the Adagrad optimizer. 

The SMCycleGAN model was developed on top of the CycleGAN source code. It inherited most of the training details in \cite{zhu2017unpaired}. The $\lambda$ that balances different components of loss functions was set to 10. Additionally, for a more stable training process, an image pool is built that stores a number of generated samples in previous iterations and is dynamically updated through training. However, due to RAM limitation, the size of the pool is restricted to 3, as opposed to 50 in \cite{zhu2017unpaired}. During the early epochs of training, the generated results are far from sensible and the method introduced in \ref{sec:masked_loss} tend not to work since the segmentation masks are poorly determined. To remedy this, a mask-out probability is implemented in a similar practice like the dropout probability. The probability to use the segmentation mask starts off to be small numbers like $0.1$, and is increased gradually as training proceeds and more realistic images are generated. The SMCycleGAN model is trained for 100 epochs.

\textbf{Baseline models.}
To demonstrate that my method indeed brings better performance, I include two baseline models as competitors. The first is the vanilla CycleGAN model \cite{zhu2017unpaired}. It is trained using the same set of data and training configurations for 100 epochs. Another baseline model is Art2Real \cite{tomei2019art2real}, which is an I2I model specialized for transforming artworks to the realistic domain. The Art2Real checkpoints are downloaded from the authors' GitHub repository and used directly for testing without further training.

\subsection{Qualitative Results}
\begin{figure*}
    \centering
    \includegraphics[width=0.8\textwidth]{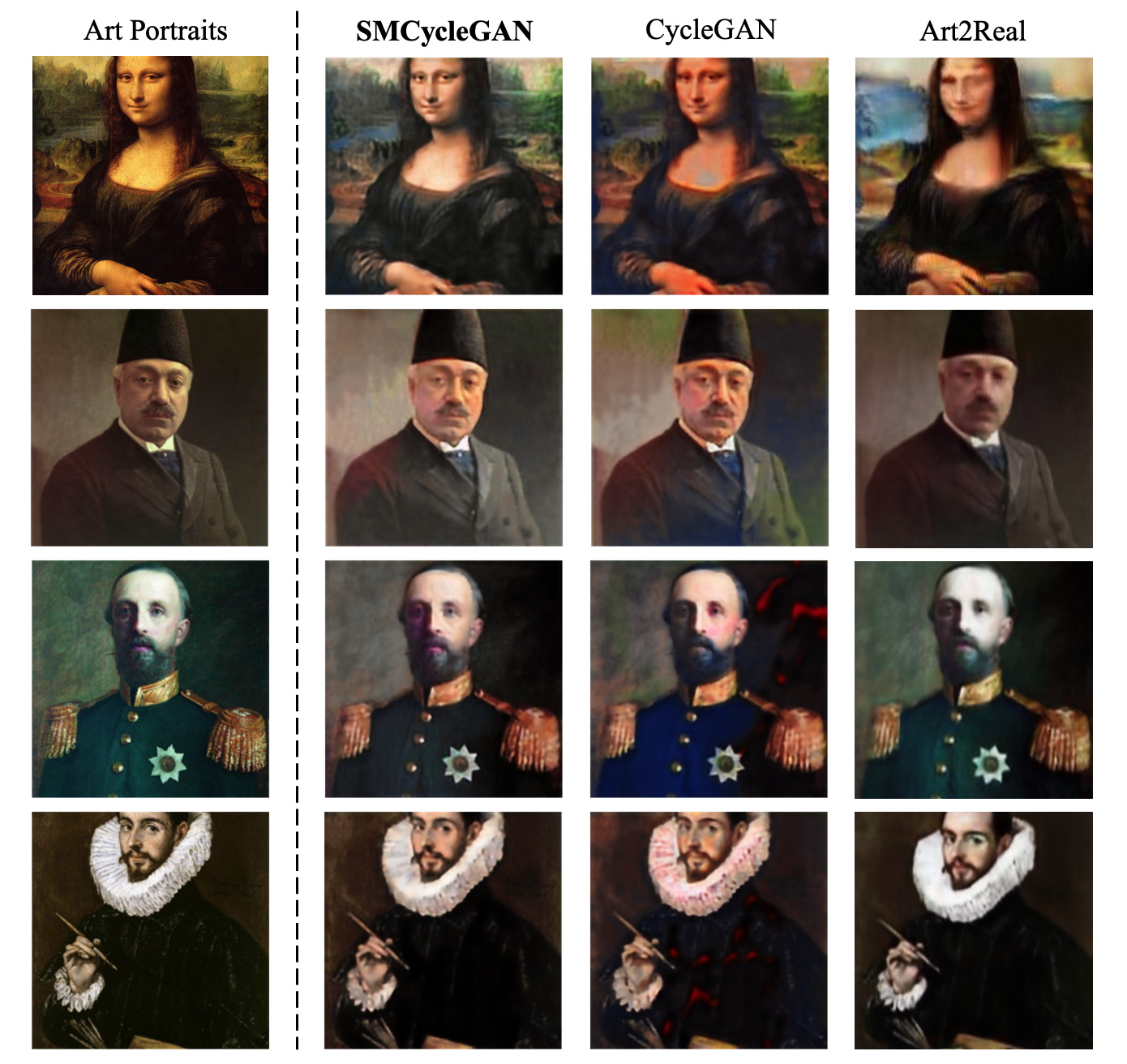}
    \caption{Qualitative results of SMCycleGAN and baseline models. My method can reduce the degree of artifacts while preserving facial details of the human subjects as opposed to CycleGAN \cite{zhu2017unpaired} and Art2Real \cite{tomei2019art2real}.}
    \label{fig:comparison}
\end{figure*}

\begin{figure*}
    \centering
    \includegraphics[width=0.8\textwidth]{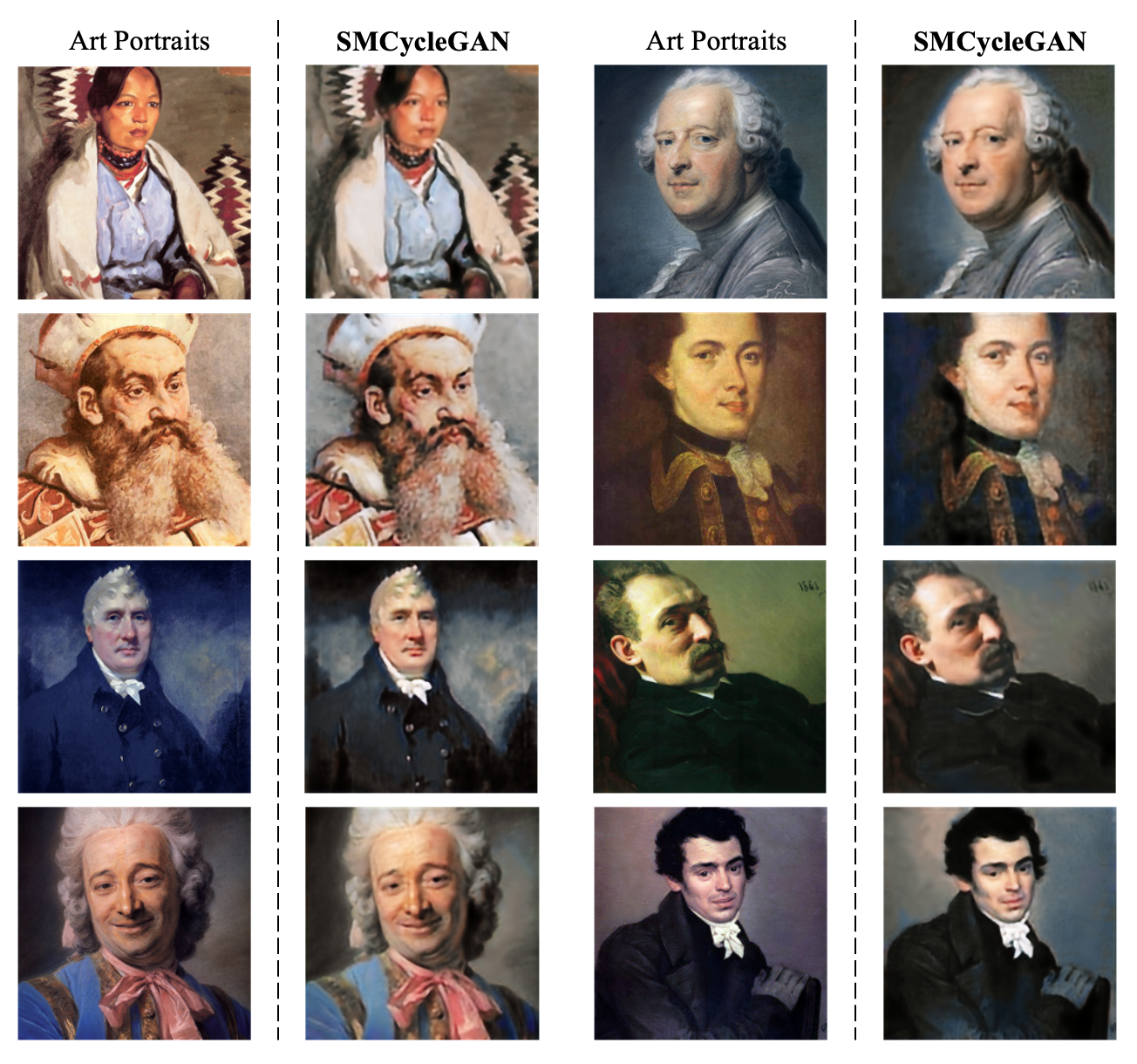}
    \caption{More relatively successful test examples of SMCycleGAN.}
    \label{fig:more}
\end{figure*}

As seen in Fig. \ref{fig:comparison}, SMCycleGAN is capable of generating portraits with high realism by adjusting skin color, smoothing texture, etc. Although the baseline models are also able to generate sensible outputs, they look less realistic due to either vagueness or anomalous color composition. In particular, the vanilla CycleGAN outputs are similar to those of my model. However, as can be observed in Fig. \ref{fig:comparison}, in some cases there are evident noises either in the background or on the subject's clothes. This is presumably caused by the mismatch of mappings mentioned in Section \ref{sec:methods}. Thus, qualitative results indicate that my method improves the vanilla CycleGAN \cite{zhu2017unpaired} by a certain degree in that in can produce cleaner and more realistic outputs. More examples are shown in Fig. \ref{fig:more}.


The Art2Real \cite{tomei2019art2real} framework relies on external image banks that store real human faces and the training aims to maximize the similarity of the generated patches and real patches. This may lead to a situation where some facial features get replaced by or shifted towards those in the image banks. For example, a subject may be assigned a new nose that belongs to someone else in the image bank. In those cases, even though the transformed images have higher realism, the faces would look different from what the subjects really looked like centuries ago, which contradicts with the original intention of this project.

\subsection{Quantitative Evaluation}
As GANs does not have an orthodox evaluation metric, it is generally much more difficult to assess their performance as opposed to models that tackle conventional vision problems such as classification, segmentation, etc. However, the Fréchet Inception Distance introduced by Heusel et al. has been universally adopted for assessment of the quality of the generated images \cite{heusel2017gans}. It measures the distance between two multidimensional Gaussian distributions. Under the setting of this project, the FID between the generated images and realistic portraits is computed. The lower FID, the more similar the two sets are. In other words, the more realistic the generated images look. 
\begin{table}[htp]
\centering

\begin{tabular}{c c} 
\hline
\multicolumn{1}{c}{Method} & \multicolumn{1}{c}{Fréchet Inception Distance} \\ \hline
Original paintings&0.6482\\
Vanilla CycleGAN \cite{zhu2017unpaired}&0.5880\\\hline
\textbf{SMCycleGAN} & \textbf{0.4912}\\\hline
\end{tabular}
\caption{Comparison in terms of the Fréchet Inception Distance \cite{heusel2017gans} of SMCycleGAN and the vanilla CycleGAN \cite{zhu2017unpaired}.}
\label{table:fid}
\end{table}

As can be seen in Table \ref{table:fid}, the SMCycleGAN model achieved the lowest Fréchet Inception Distance, meaning that it generates the most similar results to realistic portrait photos. The numerical improvement indicates that my method works robustly for translating artistic portraits to realistic photos.

\section{Conclusion}
\label{sec:conclusion}

This project have presented SMCycleGAN, a novel model for translating artistic portraits to photo-realistic visualizations. The motivation of this project comes from the wish to restore the ground truth appearances of the subjects portrayed centuries ago, with the artistic effects ruled out. The model was trained by adopting a method where semantic information is exploited. In particular, a much higher penalty is applied on the segmented regions of the objects of our interest, in my case the human subjects, while the other objects induce a relatively low level of penalty. This gives the generators more incentive to refine its performance on the human subjects generation, and less incentive to shift their attention to match the backgrounds to realistic scenarios. The model has been shown to produce compelling results. Additionally, SMCycleGAN achieved more realistic results and lower Fréchet Inception Distances \cite{heusel2017gans} compared to the vanilla CycleGAN \cite{zhu2017unpaired} and Art2Real \cite{tomei2019art2real}, which are state of the art models for I2I tasks from the artistic domain to the realistic domain.
\begin{figure}
    \centering
    \includegraphics[width=0.5\textwidth]{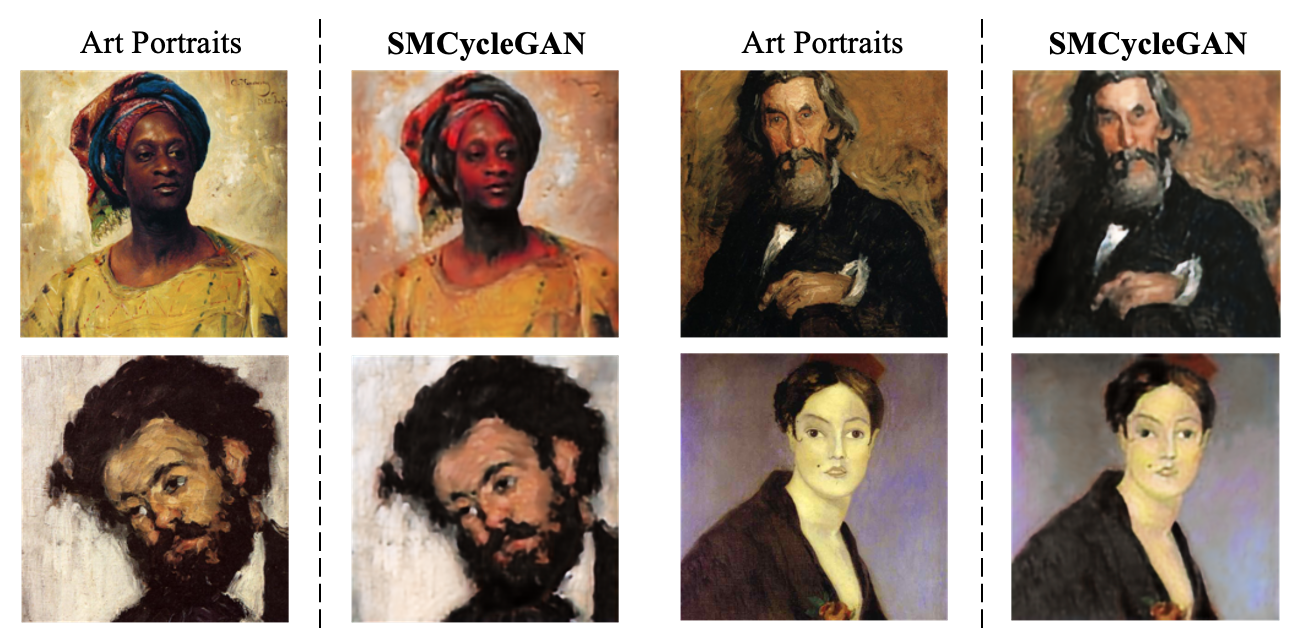}
    \caption{Failure cases of my model. Left: SMCycleGAN produces poor results for African subjects due to the small number of black people in the training set. Right: SMCycleGAN fails for artworks that are too abstract.}
    \label{fig:failure}
\end{figure}
Despite the compelling results, my model has several limitations. First, the model fails to generate realistic results for Africans or Asians. As seen in Fig. \ref{fig:failure}, the test example on the top-left is an African, and the transformation is poorly done as the generated subject has anomalous red skin. This is presumably due to the imbalanced ethnicity composition in the training set where a majority are white people. Similar cases are observed for other ethnicity such as Asians. Other failures happened when the input portrait is too abstract or contains a large amount of artifacts. As can been seen in Fig. \ref{fig:failure}, the other 3 test examples are abstract artworks that do not embed adequate information for the model to recover the subjects' appearances. Under such circumstances, the Art2Real model \cite{tomei2019art2real} may outperform SMCycleGAN as it retrieves real patches from external memory banks and this may compensate for the lack of information in the source images.

Nevertheless, the SMCycleGAN model proposed in this project provides a novel way of unsupervised image-to-image translation tasks, which exploits semantic information embedded in the source images and adjust adversarial loss accordingly. Further work can be done to explore different problems other than translating art portraits to photo-realistic images.

\textbf{Acknowledgements:} This project was developed primarily based on \href{https://www.kaggle.com/code/songseungwon/cyclegan-tutorial-from-scratch-monet-to-photo}{this Kaggle notebook} that implemented the vanilla CycleGAN \cite{zhu2017unpaired} and \href{https://www.kaggle.com/code/gamerplayer/face-matting}{this Kaggle notebook} that implemented U-Net \cite{ronneberger2015u} for human body segmentation.
{\small
\bibliographystyle{ieee_fullname}
\bibliography{reference}
}

\end{document}